\documentclass{article}

\PassOptionsToPackage{numbers,square,sort&compress}{natbib}

 \usepackage[preprint]{neurips_2026}

\workshoptitle{AI at Scale for Clinical Impact (ASCI): Cancer Pathology Foundation Models}

\usepackage[utf8]{inputenc}
\usepackage[T1]{fontenc}
\usepackage{amsmath,amssymb,mathtools,bm}
\usepackage{booktabs,tabularx,array,multirow}
\usepackage{graphicx,microtype,placeins}
\usepackage{xcolor}
\usepackage[hidelinks]{hyperref}
\usepackage{tikz}
\usetikzlibrary{arrows.meta,positioning,fit,backgrounds}
\definecolor{navy}{HTML}{173B57}
\definecolor{teal}{HTML}{187A78}
\definecolor{paleBlue}{HTML}{EAF2F8}
\definecolor{paleTeal}{HTML}{E7F4F2}
\newcommand{\method}{\textsc{SlideRuler}}
\newcommand{\R}{\mathbb{R}}
\newcommand{\med}{\operatorname*{med}}
\newcommand{\clip}{\operatorname{clip}}
\newcommand{\tablefont}{\small}
\newcolumntype{Y}{>{\raggedright\arraybackslash}X}

\author{%
  Ming Ren Hou \\
  App-In Club \\
  \texttt{ming.hou@appinclub.org} \\
  \And
  Tianyi Huang \\
  Ryquo \\
  \texttt{tianyi@ryquo.com} \\
}

\title{One-Slide Calibration of Pathology Foundation Models}

\begin{document}

\maketitle

\begin{abstract}
Scanner variation changes how pathology foundation models represent the same tissue. We introduce \method{}, which uses other regions within a slide as internal controls to predict a query region's scanner-induced shift. A transfer map learned from paired rescans enables calibration from a single scan at inference while keeping the foundation model fixed. Across two encoders and five SCORPION scanners, learned transfer reduces mean target-to-source embedding distance by 16.3--38.5\% relative to raw embeddings. Replacing controls with unrelated same-scanner regions reveals a positive same-slide contribution in all four evaluation settings, including scanner holdout. A source-anchored variant reduces source-feature displacement by 47.7--83.6\% relative to learned transfer while retaining most of its alignment gain. By drawing calibration information from the slide itself, \method{} offers a path toward more consistent use of frozen pathology models across imaging systems.
\end{abstract}

\section{Introduction}

A pathology foundation model sees both tissue and the imaging system that digitized it. Optical response, focus, compression, and color processing can change a representation even when the glass slide is fixed \citep{duenweg2023scanner,ji2025color}. Scanner robustness is therefore partly a measurement problem: how can a model distinguish acquisition variation from the biological information that its downstream tasks require? This distinction matters for deployment across hospitals and devices, where retraining a foundation model or assembling a target-domain calibration cohort may be impractical.

Paired rescans provide a direct way to measure acquisition variation. Comparing the same tissue across devices reduces the biological confounding present in comparisons between unrelated cohorts. Recent methods use this evidence to learn global transformations of frozen embeddings \citep{donle2026featmap,oreiller2026shrinkage}. An additional question remains: \emph{does the slide being processed contain information about its own correction beyond the correction shared across slides?} Answering it requires a comparison that preserves the scanner while changing the identity of the control tissue.

\method{} estimates acquisition shifts from other regions of the same slide. Paired training scans define an acquisition subspace and supervise a map from control displacement to query-region displacement. At inference, the model uses source prototypes and controls from the current scan, without requiring a paired rescan, scanner identity, or target-cohort statistics.

We study two estimators. Learned transfer combines direct control subtraction with an affine ridge map. A source-anchored variant combines a control-modulated global shift with a regularized residual, using source-feature displacement to guide model selection. Prostate biopsies motivate tissue-defined controls: we investigate benign-gland and stromal controls in synthetic transfer experiments and evaluate their selection and availability in real data.

Our contributions connect three aspects of one-slide calibration. First, same-scanner control replacement measures the additional predictive information supplied by the current slide. Second, a source-anchored residual estimator provides an explicit tradeoff between target alignment and source-feature displacement. Third, patient-held-out tissue selection and a fixed-cohort prostate-biopsy study evaluate the availability of tissue-defined controls. These analyses connect a slide's calibration signal to the correction it induces and the tissue needed to estimate it.

\section{Related work}

\paragraph{Acquisition harmonization.}
Color transfer and stain normalization address appearance variation \citep{reinhard2001color,macenko2009normalizing}; ComBat and CORAL adjust feature distributions using batch structure or domain statistics \citep{johnson2007combat,sun2016coral}. Paired-data approaches such as FEATMAP and post-hoc acquisition shrinkage learn corrections directly in foundation-model embedding spaces \citep{donle2026featmap,oreiller2026shrinkage}. PathoROB evaluates biological representation structure alongside downstream robustness, motivating evaluation of both alignment and information preservation \citep{komen2026pathorob}. Our focus is the incremental information available within one test slide. We compare against explicit pooled-mean, scanner-mean, affine-ridge, and subspace-shrinkage implementations.

\paragraph{Control tissue and scanner benchmarks.}
MuSClD uses off-target organs collected at a deployment site for cross-organ calibration \citep{zhou2023muscld}; non-tumor regions have also been used to define pseudo-domains for pathology domain generalization \citep{shigeyasu2025domain}. \method{} uses within-slide controls to predict a region-specific acquisition shift. SCORPION supplies 480 aligned regions scanned on five devices \citep{ryu2025scorpion}. PAR contains three scans of each of 339 prostate-biopsy slides from 185 patients \citep{muhammadali2026par}. Together, paired-region and whole-slide data support separate measurements of acquisition transfer and control availability.

\begin{figure}[t]
\centering
\includegraphics[width=\linewidth]{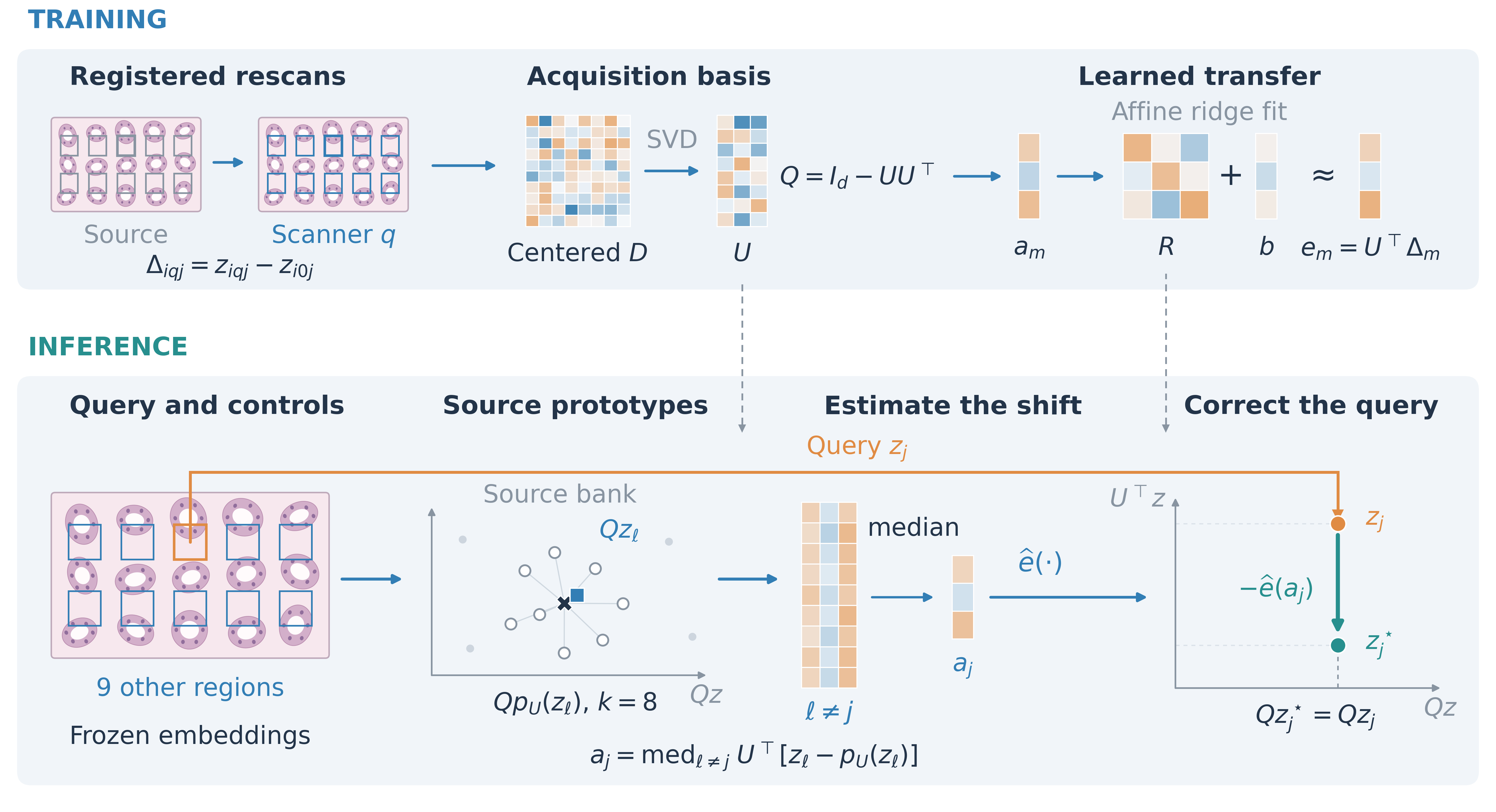}
\caption{\textbf{One-slide calibration with learned transfer.}
Top: paired rescans define the acquisition basis $U$ from residuals centered across scanners within each registered region and train the transfer map. Bottom: a single scan supplies control displacements relative to source prototypes; learned transfer corrects the query while preserving $Qz_j$. Controls exclude the query; training prototype matches exclude its entire slide. The nine-control SCORPION example is schematic.}
\label{fig:method}
\end{figure}

\section{One-slide acquisition calibration}
\label{sec:method}

\paragraph{Learned transfer in plain terms.}
Consider a SCORPION slide with ten regions: one is the query, and the other nine supply controls. Each control is matched to similar source-scanner regions from the training bank, after suppressing estimated scanner-sensitive directions. The median displacement of these controls summarizes the current scan. A map trained on paired rescans predicts the query's shift from that summary, and subtracts the prediction from the query embedding (Figure~\ref{fig:method}). Inference uses the current scan, source bank, and training-fitted parameters. Controls are reference regions; prototypes summarize their source-bank matches.

\subsection{Paired training and within-slide controls}

Let $z_{iqj}\in\R^d$ be a frozen encoder's embedding for region $j$ of slide $i$ on scanner $q$, standardized with source-training statistics. Scanner $0$ is the designated source. Registered training regions define shifts $\Delta_{iqj}=z_{iqj}-z_{i0j}$. All scans of a physical slide remain in the same fold. Training permits paired rescans; inference receives only the query scan.

Let $\mathcal Q_{\rm tr}$ contain the permitted training scanners, including source scanner $0$. For learned transfer, center each registered region across these scanners and stack the residuals:
\begin{equation}
 d_{iqj}=z_{iqj}-\frac{1}{|\mathcal Q_{\rm tr}|}
 \sum_{q'\in\mathcal Q_{\rm tr}}z_{iq'j},\qquad
 U=[u_1,\ldots,u_r],\qquad Q=I_d-UU^\top .
 \label{eq:basis}
\end{equation}
The rows of $D$ are $d_{iqj}^{\top}$, and $u_1,\ldots,u_r$ are its leading right singular vectors. Thus $U\in\R^{d\times r}$ has orthonormal columns. We retain the smallest rank reaching 90\% of residual energy, or the rank cap of 16 when that threshold is not reached. The projector $Q$ removes these acquisition coordinates for prototype matching.

Let $p_U(z)$ be the mean of the $k=8$ nearest source-bank embeddings to $z$ under distance $\|Q(z-p)\|_2$. The control vector for query region $j$ in scan $X$ is
\begin{equation}
 a_j(X;U)=\med_{\ell\ne j}U^\top\!\left[z_{\ell}-p_U(z_{\ell})\right].
 \label{eq:control}
\end{equation}
The median is coordinate-wise, and $a_j\in\R^r$. Training matches exclude every prototype from the query slide; held-out slides use the training bank. On SCORPION, each query has nine other unannotated regions as controls. The synthetic tissue-defined extension instead uses selected benign-gland and stromal fields, with separate prototype banks and transfer maps (Appendix~\ref{app:compartment}).

\subsection{Learned transfer}

Write $m$ for a training slide/scanner/query example, $a_m$ for its control vector, and $e_m=U^\top\Delta_m$. Fit an affine ridge map with $R\in\R^{r\times r}$ and $b\in\R^r$, then blend its prediction with direct subtraction:
\begin{equation}
 \begin{aligned}
 (R,b)&=\arg\min_{R,b}\sum_m\|e_m-Ra_m-b\|_2^2+\lambda\|R\|_F^2,\\
 \widehat e(a)&=(1-\alpha)a+\alpha(Ra+b),\qquad z_j^\star=z_j-U\widehat e(a_j).
 \end{aligned}
 \label{eq:learned}
\end{equation}
Source examples have $\Delta_m=0$. The original region-transfer estimator uses $\lambda=1$. Five inner physical-slide folds rebuild standardization, basis, prototypes, and regression. The shared blend $\alpha\in\{0,.01,\ldots,1\}$ minimizes a worst-scanner score based on slide-mean projected residual distance (Appendix~\ref{app:selection}). The direct-control baseline uses $\alpha=0$. The correction changes only coordinates in $U$, leaving $Qz_j$ unchanged.

\subsection{Source-anchored alternative and estimator choice}

Learned transfer is the reference estimator for measuring slide-specific scanner information. The source-anchored alternative adds movement of source-scanner features to the calibration objective. It centers the control vector at its training-source mean, uses that centered vector to modulate a pooled global shift, and adds a norm-capped ridge residual. The correction is exactly zero at the centered control vector zero, but need not be zero for every source scan. The variant also changes the basis, centering, and parameterization (Appendix~\ref{app:anchor}).

Five inner grouped folds select among 14 candidates, including identity, by the equally weighted objective
\begin{equation}
 M^\star=\arg\min_{M\in\mathcal M}
 \left[\max_{q\in\mathcal Q_{\rm tr}\setminus\{0\}}\overline{\|\Delta-\widehat\Delta_M\|_2}_{\,q}
 +\overline{\|\widehat\Delta_M\|_2}_{\,0}\right].
 \label{eq:selection}
\end{equation}
Each mean weights physical-slide groups equally. All learned components are rebuilt in every fold, and the selected candidate is refitted on outer training data. Thus learned transfer targets alignment, while the anchored alternative is appropriate when source-feature displacement is also an explicit cost. Task-specific choice between either estimator, global corrections, and identity requires downstream validation; source displacement alone cannot establish diagnostic benefit.

\subsection{Tissue-defined controls}

For prostate biopsies, we train a three-class linear selector on frozen Prost40M embeddings from an external annotated dataset \citep{calapaqui2026biopsies}. Its classes are benign gland, residual stroma, and tumor. A field is retained when its largest predicted class probability is at least 0.8; otherwise it is unassigned. The cutoff is fixed before evaluation and applies to uncalibrated probability estimates. Five outer patient-grouped folds evaluate the selector; three inner folds select logistic-regression regularization by class- and patient-balanced out-of-fold log loss. Both levels use stratified group splits. The final selector is fitted on all external patients and then frozen for PAR, where fields in each scanner image are selected independently from their uncorrected embeddings.

\section{Experimental design}
\label{sec:experiments}

\paragraph{Paired scanner evaluation.}
SCORPION contains 48 physical slides, ten registered regions per slide, and five scanners \citep{ryu2025scorpion}. We extract 384-dimensional embeddings with frozen Prost40M and DistillPath-IS16-HOpt0 \citep{prost40m2026,grisi2026routine,kaspar2026distillpath,distillpathmodel2026}. Leica AT2 is the source. Six outer splits use 32 training, eight reserved validation, and eight test slides. The calibrated protocol permits training pairs from all scanners; scanner holdout excludes the evaluated device from all fitting and selection. Each encoder/protocol condition contains 1,920 held-out target-region comparisons.

\paragraph{Comparisons and endpoints.}
We evaluate raw embeddings, pooled-mean and scanner-mean correction, affine ridge, subspace shrinkage, direct controls, learned transfer, and the anchored variant. All SCORPION alignment comparisons apply the estimators without the optional strict gate. We measure mean target-to-source Euclidean distance in source-standardized coordinates, source-feature displacement $\|z^\star-z\|_2$ on source scans, and top-1 same-region retrieval accuracy among the 80 source regions in an outer test fold. Material distance worsening means an increase exceeding 2\% of the median raw non-source distance on the eight reserved validation slides. The anchored variant does not use those slides for fitting or candidate selection.

\paragraph{Isolating slide identity.}
Thirty-two seeded derangements replace only the control regions with those of another slide on the same scanner and in the same test fold. The query embedding remains fixed. We average the paired effects across permutations before computing pointwise 95\% intervals from 5,000 physical-slide bootstrap resamples. The comparison measures the benefit of same-slide controls while holding the query, scanner, and learned estimator fixed. The donor pool and seeded assignments remain fixed during resampling.

\paragraph{External selection and PAR sampling.}
The external selector study uses 8,325 fields from 37 slides and 32 patients. Evaluation gives each patient total weight one; 95\% intervals use 2,000 patient bootstrap resamples of saved out-of-fold predictions. For PAR, a fixed patient ordering selects 20 patients and all their biopsies, yielding 35 slides and 105 scans. Reference and dense policies request 512 and 2,048 fields per scan, respectively, sampled independently on each scanner image at a nominal 112\,$\mu$m field size, with up to 128 and 512 registered fields. The policies use non-nested sampled locations and otherwise fixed extraction, selection, and registration settings. Control-count eligibility requires at least 16 benign-gland and 16 stromal fields per scan. All assigned scans remain in the denominators.

\paragraph{Controlled compartment experiment.}
A 64-run synthetic suite combines eight regimes and eight seeds, with known acquisition shifts and separate benign, stromal, and lesion compartments. It evaluates the two-compartment transfer and selective-correction implementation. All seeds contribute to the reported means; a failed outer screen returns the uncorrected embeddings. The suite tests acquisition-sensitive and already robust ordinal heads, tissue-response differences, scanner extrapolation, and calibration-set size.

\section{Results}
\label{sec:results}

\begin{figure}[t]
\centering
\includegraphics[width=\linewidth]{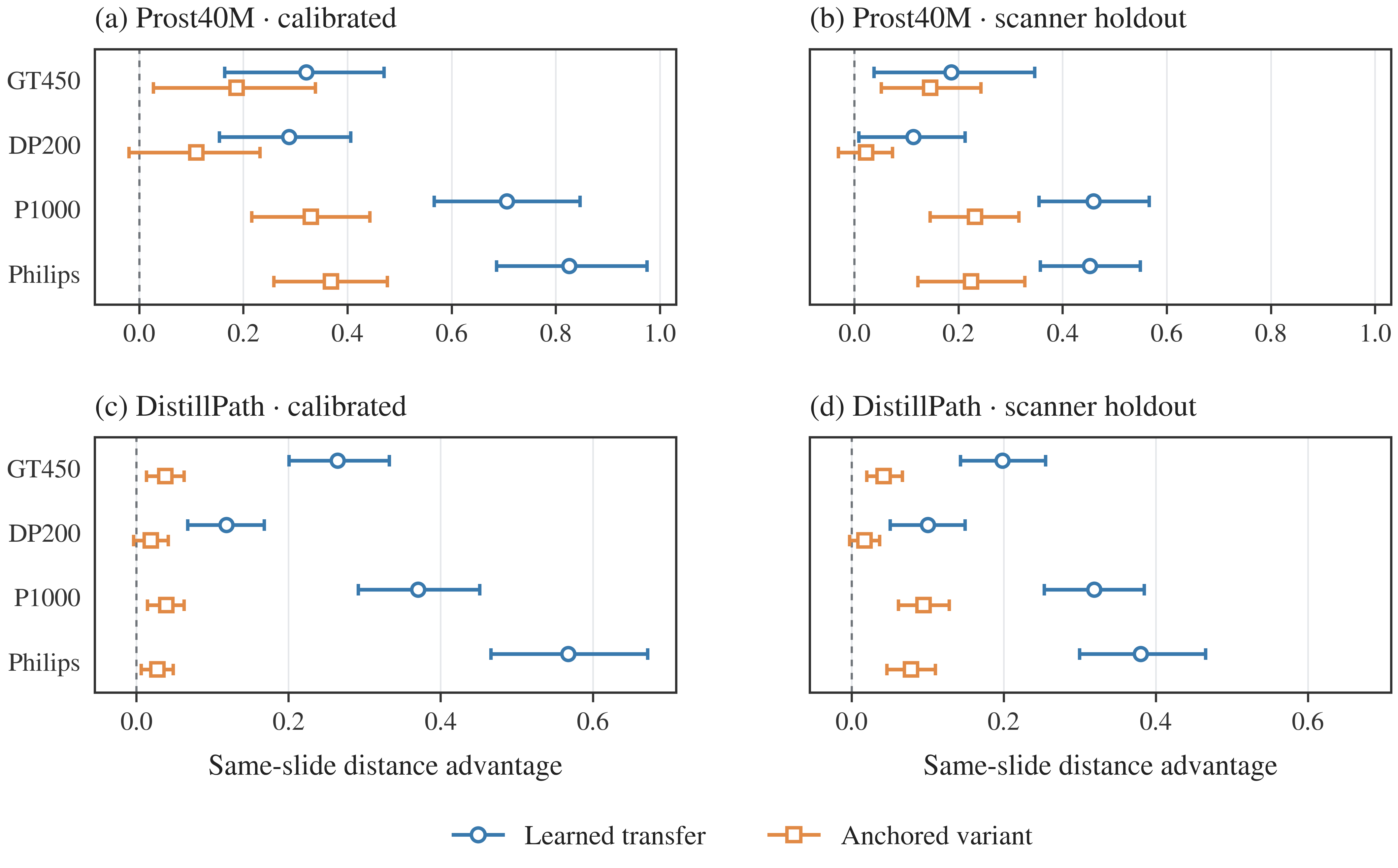}
\caption{\textbf{Slide-specific acquisition information across scanners.} Same-slide distance advantage over unrelated same-scanner controls, disaggregated by target device. Positive values favor same-slide controls. Points and bars show mean differences and pointwise 95\% physical-slide bootstrap intervals over 48 slides, after averaging 32 within-fold donor permutations. The query embedding is unchanged. Limits are shared within encoder rows; distances use each encoder's source-training scale.}
\label{fig:effects}
\end{figure}

\subsection{Same-slide controls improve acquisition correction}

Learned transfer reduces paired distance relative to raw embeddings in every encoder/protocol setting, by 16.3--38.5\% (Table~\ref{tab:alignment}). Same-slide controls improve on unrelated same-scanner controls by 0.535 and 0.303 for Prost40M, and by 0.330 and 0.249 for DistillPath, in calibrated and held-out settings respectively. All four pooled intervals exclude zero, as do all 16 learned-transfer scanner-specific intervals; the pooled effect is positive under each of the 32 permutations. Figure~\ref{fig:effects} disaggregates the effect by device. These results show that same-slide controls improve prediction of paired acquisition shifts beyond unrelated controls from the same device.

\begin{table}[ht]
\centering\tablefont
\setlength{\tabcolsep}{3.2pt}
\caption{\textbf{Paired embedding alignment on SCORPION.} Mean target-to-source distance; lower is better. All corrections are ungated. Bold denotes the lowest point estimate in each row. Scanner mean falls back to pooled mean for a held-out scanner. Distances use each encoder's source-training scale.}
\label{tab:alignment}
\begin{tabular*}{\linewidth}{@{\extracolsep{\fill}}llrrrrrr@{}}
\toprule
Encoder & Protocol & Raw & Scanner mean & Affine ridge & Shrinkage & Learned & Anchored\\
\midrule
Prost40M & Calibrated &18.654&11.328&\textbf{9.640}&13.181&11.470&12.030\\
& Holdout &18.654&15.940&\textbf{13.294}&14.801&13.871&14.196\\
DistillPath & Calibrated &17.551&\textbf{13.805}&14.447&13.808&13.843&14.591\\
& Holdout &17.551&15.840&15.420&14.690&\textbf{14.682}&14.973\\
\bottomrule
\end{tabular*}
\end{table}

The same-slide increment accounts for 6.3--8.9\% of learned transfer's total distance reduction over raw embeddings; most of the gain remains with unrelated same-scanner controls (Appendix~\ref{app:scorpion}). Global baselines remain competitive: affine ridge has the lowest mean distance for Prost40M in both protocols, while paired distance-advantage intervals include zero versus scanner mean in calibrated DistillPath and versus shrinkage under DistillPath holdout. The contribution is therefore a consistent incremental slide signal. Retrieval and distance-worsening rates provide complementary comparisons. Under Prost40M scanner holdout, learned transfer achieves 99.74\% retrieval accuracy versus 98.18\% for affine ridge, a paired advantage of 1.56 percentage points (95\% CI 0.83--2.34), with material distance worsening of 3.07\% versus 19.58\%. Under DistillPath holdout, learned transfer and shrinkage have similar mean distance, with worsening rates of 2.40\% and 6.77\%, respectively.

\subsection{Source anchoring changes the operating point}

\begin{figure}[t]
\centering
\includegraphics[width=\linewidth]{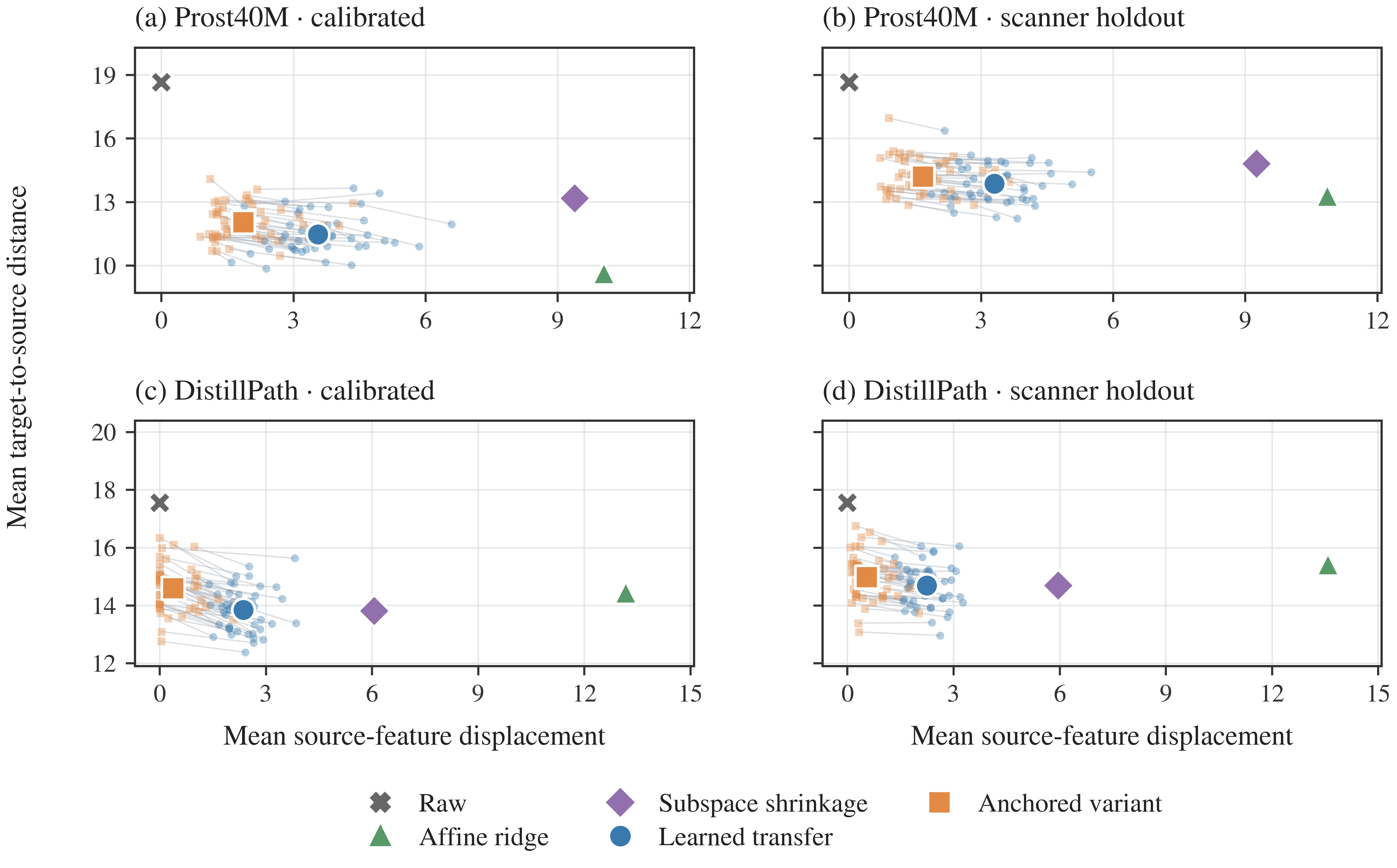}
\caption{\textbf{Alignment and source displacement form distinct evaluation axes.} Large markers are method means; faint blue/orange points show learned/anchored results for each of 48 physical slides, joined within slide. Lower values on both axes are preferable. Target values average ten regions and four target scanners; source values average ten regions and one fitted model per slide when calibrated, or four models under scanner holdout. Axis limits are shared within encoder rows; distances use source-standardized coordinates.}
\label{fig:tradeoff}
\end{figure}

The anchored estimator reduces mean source-feature displacement by 47.7--83.6\% across the four settings. Figure~\ref{fig:tradeoff} shows the joint alignment--displacement comparison and its physical-slide variation; source displacement decreases on all 48 slides in every setting. This reduction accompanies an increase of 0.291--0.749 in mean target distance relative to learned transfer; all four paired intervals exclude zero. Pooled material distance-worsening rates also decrease in every setting (Appendix~\ref{app:residual}). Anchoring retains 79.8--93.2\% of learned transfer's distance reduction over raw embeddings, providing a way to reduce source displacement while preserving most of the target alignment gain.

Training-only model selection retains a residual component in 42 of 60 unique fits and selects the control-modulated global correction in the remaining 18. All Prost40M fits retain a residual, whereas all six calibrated DistillPath fits select the global correction. This variation shows how model selection adjusts correction complexity across encoders. The anchored estimator retains positive pooled same-slide advantages of 0.031--0.249 (Appendix~\ref{app:scorpion}).

\begin{figure}[t]
\centering
\includegraphics[width=\linewidth]{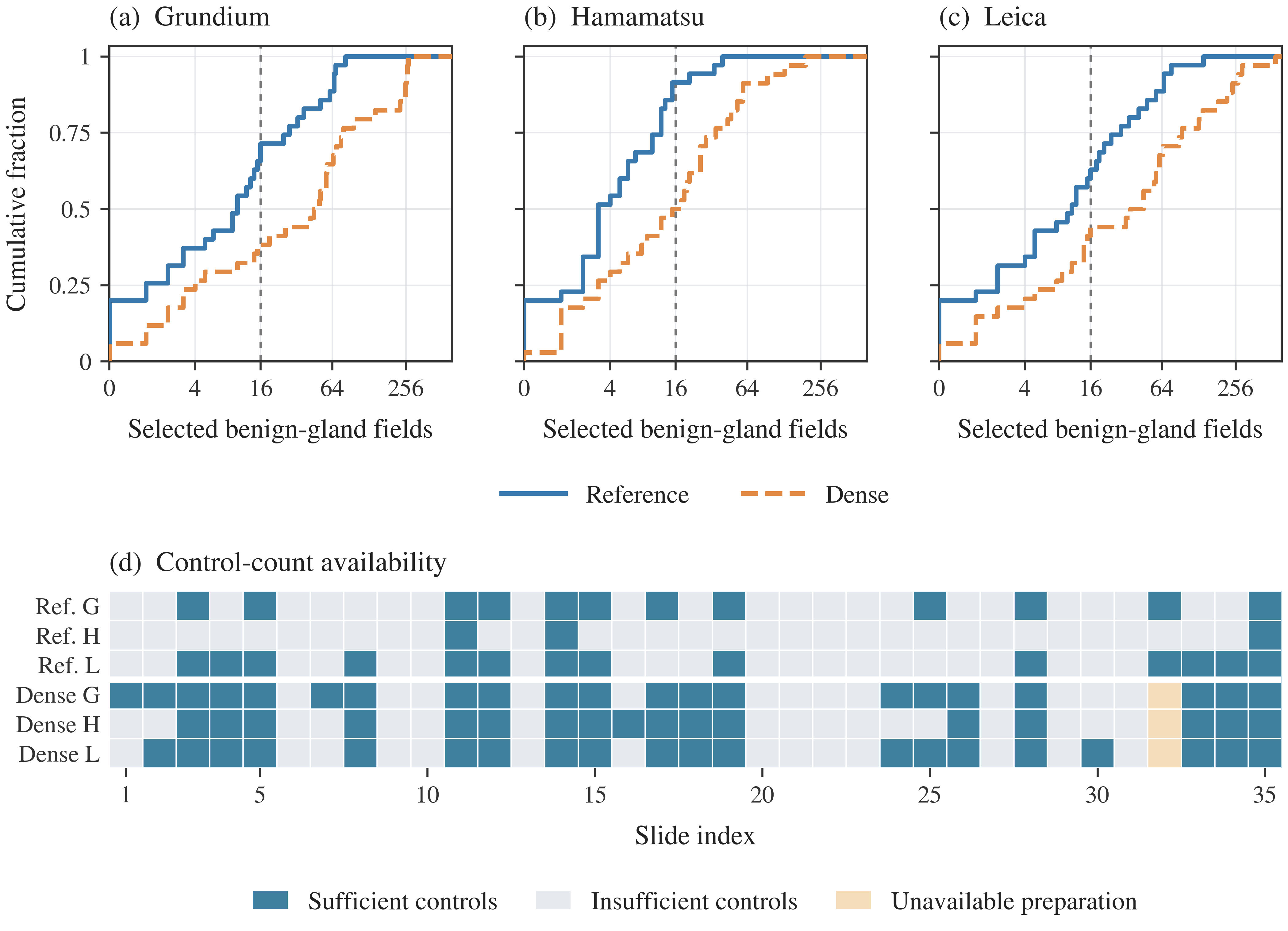}
\caption{\textbf{Control availability across sampling policies.} (a--c) Empirical cumulative distributions of benign-gland counts for 35 reference and 34 dense scans per scanner. The horizontal coordinate is $\log_2(n+1)$, with ticks labeled by count $n$; dashed lines mark 16 fields. (d) Availability requires at least 16 benign-gland and 16 stromal fields per scan. Columns represent all 35 slides, with unavailable preparations marked. G/H/L denote Grundium/Hamamatsu/Leica. Sampling locations are non-nested.}
\label{fig:control-availability}
\end{figure}

\subsection{Denser sampling increases tissue-control availability}

\begin{table}[ht]
\centering\tablefont
\caption{\textbf{External patient-held-out tissue selection.} Precision and recall are percentages with 95\% patient bootstrap intervals. Counts denote reference-class fields and patients containing that class. Residual stroma is the dataset's postprocessed remaining-tissue class.}
\label{tab:selector}
\begin{tabular*}{\linewidth}{@{\extracolsep{\fill}}lrrcc@{}}
\toprule
Reference class & Fields & Patients & Precision & Recall\\
\midrule
Benign gland &337&21&89.37 [80.05, 95.72]&89.51 [82.19, 94.40]\\
Residual stroma &7,028&32&99.98 [99.93, 100.00]&98.16 [96.78, 99.18]\\
Tumor &960&20&96.32 [91.85, 98.69]&81.40 [71.34, 89.38]\\
\bottomrule
\end{tabular*}
\end{table}

The external selector achieves patient-weighted precision of 89.37\% for benign glands, 99.98\% for residual stroma, and 96.32\% for tumor (Table~\ref{tab:selector}). Coverage is 96.75\%. Four additional patient partitions give benign precision of 90.81--93.05\% at the same cutoff.

On PAR, dense sampling doubles the number of scans meeting both control-count requirements from 29/105 to 58/105 (Table~\ref{tab:par}). Among the 102 scans available under both policies, 27 pass under both, 31 pass only under dense sampling, and none pass only under reference sampling. Hamamatsu shows the largest absolute increase, from 3 to 17 eligible scans out of 35. Every available dense scan meets the stromal count threshold, identifying benign-gland sampling as the remaining constraint on control availability. The full distributions and paired scanner outcomes appear in Figure~\ref{fig:control-availability}. At slide level, all three scanners meet the control requirements on 3/35 reference slides and 16/35 dense slides.

\begin{table}[ht]
\centering\tablefont
\caption{\textbf{PAR control-count availability.} Scans with at least 16 selected benign-gland and 16 stromal fields. Each scanner has 35 assigned scans. The three unavailable dense scans remain in the denominator.}
\label{tab:par}
\begin{tabular*}{\linewidth}{@{\extracolsep{\fill}}lrrrr@{}}
\toprule
Policy & Grundium & Hamamatsu & Leica & Overall\\
\midrule
Reference (512 fields) &12/35&3/35&14/35 & 29/105 (27.6\%)\\
Dense (2,048 fields) &21/35&17/35&20/35 & 58/105 (55.2\%)\\
\bottomrule
\end{tabular*}
\end{table}

\subsection{Controlled tests explain when transfer helps}

In the acquisition-sensitive synthetic regime, the two-compartment method reduces shift root-mean-square error (RMSE) from 4.014 to 1.894, with paired reduction 2.120 (95\% CI 1.468--2.512), and increases worst-target quadratically weighted kappa (QWK) from 0.828 to 0.949. Averaged over all eight seeds, grade flips fall from 56.7\% to 15.7\%. Source embeddings remain unchanged in this regime. The corresponding kappa gain is smaller for an already robust head, 0.009 versus 0.121. When lesion and control responses differ strongly, learned ungated transfer improves RMSE from 3.766 for direct subtraction to 1.662. These controlled comparisons link transfer benefit to tissue-response differences and the downstream head's sensitivity to scanner shifts.

\section{Scope and limitations}

Our real-data evidence concerns embedding calibration and tissue-control availability. Both scanner datasets rescan fixed, already prepared physical slides. This isolates scanner effects but does not test independent staining, sectioning, or slide-preparation variation. SCORPION provides aligned regions and physical-slide groups, but no patient identifiers, compartment annotations, or clinical grades. Its bootstrap intervals condition on the fitted models, fixed donor pool, and development choices; public metadata also do not permit a complete audit of pretraining overlap. Control replacement changes tissue composition as well as slide identity, so its advantage measures slide-associated predictive information rather than identifying a purely technical mechanism. Estimated acquisition coordinates may contain biological variation; neither smaller paired distance nor reduced source-feature displacement establishes preservation of downstream predictions. The optional strict gate accepts no SCORPION corrections at this sample size; the reported real-scanner results concern the ungated estimators.

PAR control counts are an intermediate endpoint. Registration and matched-tumor requirements leave at most six eligible training patients, below the fixed ten-patient minimum, so no end-to-end PAR calibration or grading comparison was run. The study does not establish that benign/stromal controls calibrate tumor representations in real tissue. Control purity on PAR requires expert validation, and denser sampling yields spatially clustered fields. Synthetic stress tests show that agreement between control compartments does not protect against shared control drift or altered lesion responses. The next validation priorities are expert-verified controls, patient-separated downstream comparisons against identity and global corrections, and independent preparation shifts. The method is intended for research use; its statistical screens do not provide clinical risk guarantees.

\begin{samepage}
\section{Conclusion}

Within-slide regions provide additional information for correcting scanner-induced variation in frozen pathology embeddings. \mbox{\method{}} learns to use this signal without retraining the encoder, while the source-anchored alternative makes the alignment--displacement tradeoff explicit. Tissue selection and sampling connect the proposed tissue-defined extension to control availability in prostate biopsies. Together, these findings support within-slide calibration as a complementary approach to scanner robustness.
\par
\end{samepage}

\bibliographystyle{plainnat}
\bibliography{references}

\clearpage
\appendix

\section{Estimator and model-selection details}
\label{app:selection}

\paragraph{Original region-transfer estimator.}
The affine ridge in Equation~\ref{eq:learned} includes an unpenalized intercept. Its source examples use zero paired shift. Inner-fold transfer errors are projected Euclidean residuals $\|U^\top\Delta-\widehat e\|_2$, averaged over regions within slide. For each permitted non-source scanner $q$, write the slide-mean error as $\bar E_q$ with sample standard deviation $s_q$ over $n_q$ slides. The training selection score is
\[
 \max_q\left(\bar E_q+1.6448536\,s_q/\sqrt{n_q}\right),
\]
with ties resolved by mean scanner error, worst-scanner material-worsening fraction, and smaller blend weight. Standardization and all fitted geometric components are rebuilt in each inner fold.

\paragraph{Anchored estimator.}
Let $G$ be the number of independent training groups, $D$ the number of permitted domains including source, and $J$ the number of regions per slide. Residual-regression weights give each group/domain equal total mass and sum to $GDJ$. This fixes the normalization of the ridge penalties. For groups with several biopsies, each biopsy receives equal weight within its group; prototype matching retains all source regions. The source mean and scale, source-referenced basis, prototype bank, mean source-control anchor, pooled shift, and residual map are refitted in every inner split. Exact ties follow a fixed candidate ordering: identity, anchored global-only, then residual candidates in penalty-major order. The global shift is full-dimensional; only the residual is constrained to the acquisition subspace. The basis and source centering therefore differ from the original estimator.

\subsection{Source-anchored estimator definition}\label{app:anchor}

The second estimator balances target alignment against source-feature displacement. Its basis $U_A$ is fitted to group-weighted, \emph{source-referenced} shifts $\Delta$, and $g=\mathbb E_w[\Delta]\in\R^d$ is their pooled mean over non-source scanners. Each training group receives equal total mass. Define the centered control vector and global activation by
\begin{equation}
 \widetilde a_j=a_j(X;U_A)-\mu_0,\qquad v=U_A^\top g,\qquad
 t(a)=\clip\!\left(\frac{a^\top v}{v^\top v},0,2\right),
 \label{eq:anchor}
\end{equation}
where $\mu_0$ is the group-weighted mean of source control vectors computed with own-group prototypes excluded. Set $t=0$ if $v^\top v\le10^{-8}$. The global prediction $t(a)g$ retains the full embedding dimension.

A zero-intercept ridge map predicts the remaining acquisition coordinates:
\begin{align}
 B_\lambda&=\arg\min_B\sum_m w_m
 \left\|U_A^\top\!\left[\Delta_m-t(\widetilde a_m)g\right]-B\widetilde a_m\right\|_2^2
 +\lambda\|B\|_F^2,\label{eq:residualfit}\\
 r(a)&=\beta B_\lambda a\,
 \min\!\left\{1,\frac{\|a\|_2}{\max(\|\beta B_\lambda a\|_2,10^{-8})}\right\},
 \qquad \widehat\Delta_A(a)=t(a)g+U_A r(a).
 \label{eq:residual}
\end{align}
The fitted correction is $z_j^\star=z_j-\widehat\Delta_A(\widetilde a_j)$. Source examples again have zero target shift. The residual cap bounds its acquisition-coordinate norm by the control norm. At the learned anchor $\widetilde a=0$, the correction is exactly zero.

Four penalties $\lambda\in\{.1,1,10,100\}$ and three residual weights $\beta\in\{.25,.5,1\}$, plus global-only and identity, give 14 candidates. Equation~\ref{eq:selection} selects among them. Each mean averages regions within a slide, slides within a group, and then groups equally. Every learned component, including $\mu_0$ and $g$, is rebuilt inside each fold. The selected candidate is refitted on the outer training set. The global-only candidate retains control-dependent activation. Unlike learned transfer, the full-dimensional global correction need not preserve $Q_Az$, where $Q_A=I_d-U_AU_A^\top$.

\paragraph{Global comparators.}
Pooled mean subtracts the mean paired shift over permitted non-source training scanners. Scanner mean uses a target scanner's own training pairs when available and otherwise returns pooled mean. Affine ridge learns a regularized map from a query embedding to its source counterpart. Subspace shrinkage attenuates coordinates in a paired acquisition basis. Their hyperparameters are selected within training-only grouped folds. Paired source test regions provide evaluation references; all corrections use training-derived parameters and the query scan.

\section{Complete scanner comparisons}
\label{app:scorpion}

The same-slide increment accounts for 6.3--8.9\% of learned transfer's total distance reduction over raw embeddings. Unrelated same-scanner controls retain the remaining gain, separating the shared acquisition correction from the contribution of the current slide.

\begin{table}[ht]
\centering\tablefont
\caption{Additional baseline distances, completing Table~\ref{tab:alignment}.}
\begin{tabular*}{\linewidth}{@{\extracolsep{\fill}}llrr@{}}
\toprule Encoder & Protocol & Pooled mean & Direct controls\\\midrule
Prost40M&Calibrated&14.131&12.215\\
&Holdout&15.940&13.837\\
DistillPath&Calibrated&14.985&14.051\\
&Holdout&15.840&14.770\\\bottomrule
\end{tabular*}
\end{table}

\begin{table}[ht]
\centering\tablefont
\setlength{\tabcolsep}{3pt}
\caption{Learned transfer compared with global alternatives. Positive paired distance reductions favor learned transfer. Intervals use 5,000 paired physical-slide bootstrap resamples.}
\begin{tabular*}{\linewidth}{@{\extracolsep{\fill}}llccc@{}}
\toprule Encoder&Protocol&Versus affine ridge&Versus scanner mean&Versus shrinkage\\\midrule
Prost40M&Calibrated&$-1.830$ [$-2.007,-1.649$]&$-.142$ [$-.251,-.035$]&1.711 [1.565,1.865]\\
&Holdout&$-.576$ [$-.739,-.411$]&2.069 [1.979,2.158]&.930 [.792,1.082]\\
DistillPath&Calibrated&.604 [.411,.798]&$-.038$ [$-.091,.018$]&$-.035$ [$-.121,.052$]\\
&Holdout&.738 [.553,.922]&1.158 [1.101,1.219]&.008 [$-.064,.079$]\\\bottomrule
\end{tabular*}
\end{table}

\begin{table}[ht]
\centering\tablefont
\caption{Same-slide distance advantage over 32 wrong-slide assignments. Permutations are averaged before physical-slide resampling; the gain range describes variation across the 32 assignments.}
\begin{tabular*}{\linewidth}{@{\extracolsep{\fill}}llccc@{}}
\toprule Encoder&Protocol&Learned, 95\% CI&Anchored, 95\% CI&Learned gain range\\\midrule
Prost40M&Calibrated&.535 [.427,.641]&.249 [.169,.325]&.472--.591\\
&Holdout&.303 [.238,.369]&.156 [.099,.213]&.261--.348\\
DistillPath&Calibrated&.330 [.270,.392]&.031 [.015,.047]&.298--.367\\
&Holdout&.249 [.201,.299]&.058 [.041,.075]&.220--.277\\\bottomrule
\end{tabular*}
\end{table}

\begin{table}[!htbp]
\centering\tablefont
\caption{Top-1 retrieval accuracy (\%) and material distance worsening (\%). The worsening threshold is defined in Section~\ref{sec:experiments}.}
\begin{tabular*}{\linewidth}{@{\extracolsep{\fill}}llrrrrr@{}}
\toprule &&\multicolumn{3}{c}{Top-1 retrieval accuracy}&\multicolumn{2}{c}{Distance worsening}\\
\cmidrule(lr){3-5}\cmidrule(lr){6-7}
Encoder&Protocol&Raw&Learned&Anchored&Learned&Anchored\\\midrule
Prost40M&Calibrated&93.91&99.74&99.84&1.77&1.15\\
&Holdout&93.91&99.74&99.58&3.07&2.45\\
DistillPath&Calibrated&99.79&100.00&99.95&1.77&1.35\\
&Holdout&99.79&99.95&99.90&2.40&1.35\\\bottomrule
\end{tabular*}
\end{table}

\FloatBarrier
\section{Source-anchored refinement}
\label{app:residual}

\begin{table}[ht]
\centering\tablefont
\setlength{\tabcolsep}{3pt}
\caption{Paired anchored-minus-learned changes. Positive target-distance change means less alignment; negative worsening-rate change is favorable. Rate changes are percentage points.}
\begin{tabular*}{\linewidth}{@{\extracolsep{\fill}}llccc@{}}
\toprule Encoder&Protocol&Target distance, 95\% CI&Retrieval change, 95\% CI&Worsening change, 95\% CI\\\midrule
Prost40M&Calibrated&.560 [.466,.652]&.104 [.000,.260]&$-.625$ [$-1.094,-.260$]\\
&Holdout&.326 [.264,.389]&$-.156$ [$-.521,.156$]&$-.625$ [$-1.302,.000$]\\
DistillPath&Calibrated&.749 [.677,.822]&$-.052$ [$-.156,.000$]&$-.417$ [$-.833,-.052$]\\
&Holdout&.291 [.241,.344]&$-.052$ [$-.156,.000$]&$-1.042$ [$-1.562,-.573$]\\\bottomrule
\end{tabular*}
\end{table}

\begin{table}[ht]
\centering\tablefont
\caption{Source-feature displacement and model selection. Counts refer to six unique fits per calibrated setting and 24 per holdout setting. Source means average 48 physical slides, each evaluated under one calibrated model or four holdout models.}
\begin{tabular*}{\linewidth}{@{\extracolsep{\fill}}llrrrrr@{}}
\toprule &&\multicolumn{2}{c}{Source displacement}&\multicolumn{3}{c}{Selected models}\\
\cmidrule(lr){3-4}\cmidrule(lr){5-7}
Encoder&Protocol&Learned&Anchored&Global-only&Residual&Identity\\\midrule
Prost40M&Calibrated&3.562&1.861&0&6&0\\
&Holdout&3.306&1.683&0&24&0\\
DistillPath&Calibrated&2.365&.388&6&0&0\\
&Holdout&2.248&.555&12&12&0\\\bottomrule
\end{tabular*}
\end{table}

Every selected residual uses ridge penalty 100 and weight 0.25 or 0.5. The source-displacement means and physical-slide distributions in Figure~\ref{fig:tradeoff} use the same evaluation units.

\section{Tissue selection and PAR preparation}
\label{app:par}

\begin{figure}[t]
\centering
\includegraphics[width=\linewidth]{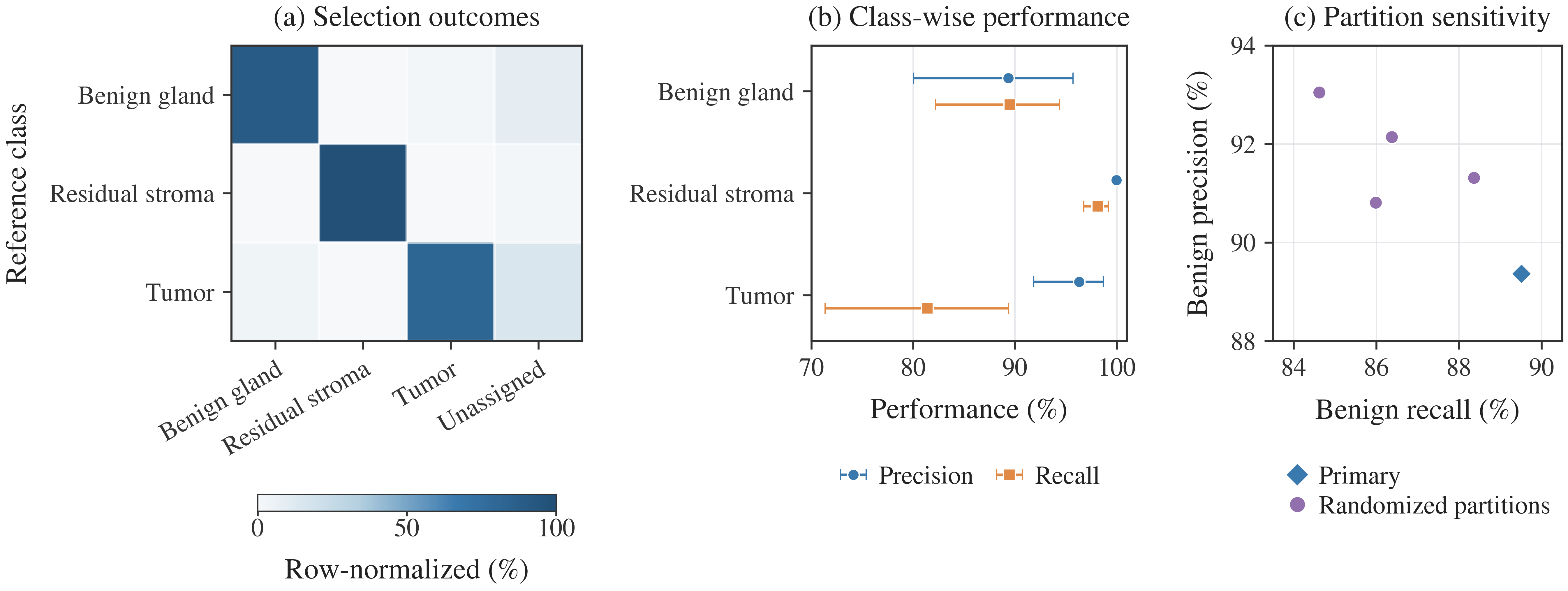}
\caption{\textbf{External tissue-selector validation.} Each patient's complete field set has total weight one. (a) Row-normalized primary out-of-fold selection outcomes, including unassigned fields at the fixed 0.8 cutoff. (b) Precision and recall with 95\% intervals from 2,000 patient bootstrap resamples; recall includes unassigned reference fields. (c) Benign precision and recall for the primary partition and all four post-result randomized partitions of the same 32 patients. The four additional partitions assess sensitivity to patient allocation.}
\label{fig:selector-validation}
\end{figure}

\paragraph{External annotations and fitting.}
The external release contains 37 slides from 32 patients \citep{calapaqui2026biopsies}. Benign labels cover whole glands, including lumen; the residual-stroma class is assigned by postprocessing remaining tissue. The selector uses 224-pixel inputs at 0.5\,$\mu$m/pixel and adapts the native crop to each image's recorded resolution. Training balances classes and patients within class; evaluation weights each patient's complete field set to total one. Inner folds choose the inverse regularization strength $C\in\{.01,.1,1,10\}$ by minimizing out-of-fold log loss with equal class weight and equal patient weight within class. Scaling and all fitting are restricted to training patients.

All five primary outer folds select $C=1$. Among 327 out-of-fold benign selections, 301 have benign reference labels, 24 tumor labels, and two residual-stroma labels. Patient-weighted accuracy, counting unassigned fields as errors, is 95.96\%. Four additional randomized patient partitions, specified after the primary results, assess sensitivity to patient allocation at the same cutoff. Their benign precision is 90.81--93.05\% and recall is 84.62--88.37\%. PAR uses the selector fitted under the primary protocol.

\paragraph{Cohort and acquisition.}
The PAR cohort uses a deterministic patient ordering, excludes the two patients used to develop tissue-compartment extraction, and retains all biopsies and all three scanners for the selected 20 patients. Cohort selection uses no grading outcomes. Reference and dense sampling share the encoder, selector, cutoff, physical scale, and seed, with non-nested sampled locations. Thumbnail tissue fraction must be at least 0.55. Reference preparations are available for all 35 slides and dense preparations for 34; the unavailable dense preparation remains in all assigned-cohort denominators.

\paragraph{Registration and joint eligibility.}
Registration uses 448\,$\mu$m context, 12 local probes, normalized cross-correlation at least 0.45, maximum local shift 64\,$\mu$m, common support at least 0.5, and cycle-consistency median/95th-percentile limits of 96/192\,$\mu$m. Automated registration quality control approves 22 reference trios and 21 dense trios, with 13 rejected under each policy and one dense preparation unavailable. The 34 matched-policy slides have identical decisions. Among approved dense trios, four Grundium--Hamamatsu and eight each Grundium--Leica and Hamamatsu--Leica pairs lack four jointly selected tumor fields. These restrictions can overlap with missing controls.

\begin{table}[ht]
\centering\tablefont
\caption{Direction-specific calibration eligibility. Each fold assigns 16 training and four held-out patients. Eligibility requires both scans' controls, approved registration, and four matched tumor fields. G: Grundium; H: Hamamatsu; L: Leica.}
\begin{tabular*}{\linewidth}{@{\extracolsep{\fill}}llrrl@{}}
\toprule Policy&Direction&Slides /35&Patients /20&Training patients, folds 0--4\\\midrule
Reference&G$\to$H&0&0&0, 0, 0, 0, 0\\
&G$\to$L&2&2&2, 2, 0, 2, 2\\
&H$\to$G&0&0&0, 0, 0, 0, 0\\
&H$\to$L&0&0&0, 0, 0, 0, 0\\
&L$\to$G&2&2&2, 2, 0, 2, 2\\
&L$\to$H&0&0&0, 0, 0, 0, 0\\\midrule
Dense&G$\to$H&5&4&3, 3, 2, 4, 4\\
&G$\to$L&5&5&4, 4, 3, 4, 5\\
&H$\to$G&5&4&3, 3, 2, 4, 4\\
&H$\to$L&3&3&2, 2, 2, 3, 3\\
&L$\to$G&5&5&4, 4, 3, 4, 5\\
&L$\to$H&3&3&2, 2, 2, 3, 3\\\bottomrule
\end{tabular*}
\end{table}

Pooling at least one eligible target direction gives reference patient totals G/H/L of 2/0/2 and dense totals of 6/4/5. Dense pooled training counts in folds 0--4 are G: 5,5,3,5,6; H: 3,3,2,4,4; L: 4,4,3,4,5. Multiple biopsies from one patient contribute one patient to these counts.

\begin{table}[ht]
\centering\tablefont
\caption{Spatial clustering of selected controls. A nearby center is another selected center of the same class within 112\,$\mu$m. The fraction measures spatial clustering of sampled fields.}
\begin{tabular*}{\linewidth}{@{\extracolsep{\fill}}llrrr@{}}
\toprule Policy&Class&Selected fields&With nearby center&Fraction\\\midrule
Reference&Benign gland&1,667&459&27.5\%\\
&Residual stroma&25,729&12,213&47.5\%\\
Dense&Benign gland&6,251&4,016&64.2\%\\
&Residual stroma&100,660&86,688&86.1\%\\\bottomrule
\end{tabular*}
\end{table}

\FloatBarrier
\section{Two-compartment selective estimator}
\label{app:compartment}

The synthetic study uses a compartment-specific instantiation. For controls $c\in\{b,s\}$, compute $a_c$ by taking the coordinate-wise median over selected control tiles using a compartment-specific source bank. Subtract the coordinate-wise median source-control vector $\mu_c$ to obtain $d_c=a_c-\mu_c$. The lesion target $e_m$ is the coordinate-wise median of projected registered lesion shifts. Two zero-intercept ridge maps predict $e_m$:
\[
 R_c=\arg\min_R\sum_m\|e_m-Rd_{cm}\|_2^2+\lambda\|R\|_F^2,
 \qquad \widehat e=\frac12\sum_{c\in\{b,s\}}
 \big[(1-\alpha)d_c+\alpha R_cd_c\big].
\]
The basis is constructed by centering each registered region across the permitted scanners, as in Equation~\ref{eq:basis}. This experiment uses ridge penalty $\lambda=0.5$ and 21 blend weights $\alpha\in\{0,.05,\ldots,1\}$. Patient-grouped out-of-fold selection minimizes non-source mean residual norm, then any-worsening rate, then blend weight. The two compartments, source-median centering, and zero-intercept maps define the instantiation used in these synthetic experiments.

Write $\widehat e_c=(1-\alpha)d_c+\alpha R_cd_c$ for each compartment prediction. Selective correction requires sufficient controls; bounded dispersion and prototype-matching radius; supported joint control coordinates under Ledoit--Wolf covariance shrinkage \citep{ledoit2004well}; and bounded disagreement
\[
 \rho=\frac{\|\widehat e_b-\widehat e_s\|_2}
 {\|\widehat e_b\|_2+\|\widehat e_s\|_2+\epsilon_\rho}.
\]
The denominator stabilizer is 0.1 times the median non-source target-shift norm, with a numerical lower bound. Prediction norms must exceed the maximum source prediction norm, computed out of fold during threshold selection and recomputed after refitting the transfer maps. Candidate thresholds maximize eligible non-source coverage subject to grouped gain and event-rate screens. An outer patient cross-fit repeats the complete procedure from supplied embeddings. If it fails, the fitted model returns identity for every slide; otherwise rejected slides return identity individually. Accepted candidate lesion tiles receive $z_j^\star=z_j-U\widehat e$.

For gate selection, material worsening is an increase in projected residual norm exceeding 2\% of the fitting set's median non-source projected target norm. For binary group indicators $H_i$, let $n$ be the number of evaluated groups and $\widehat p=n^{-1}\sum_iH_i$. The one-sided 95\% Wilson upper bound is \citep{wilson1927probable}
\begin{equation}
 U_{\rm W}(\widehat p,n)=
 \frac{\widehat p+z^2/(2n)+z\sqrt{\widehat p(1-\widehat p)/n+z^2/(4n^2)}}{1+z^2/n},
 \qquad z=1.6448536.
\end{equation}
The worsening screen uses patients with at least one accepted target scan; the source-correction screen uses all control-count-eligible source patients. The screens require the Wilson upper bounds for patient-any material worsening and patient-any source correction to be at most 5\% and 2\%, respectively, alongside nonnegative grouped lower alignment gain. Pooled and supplied domain-specific constraints are checked. At zero events, $U_{\rm W}=z^2/(n+z^2)$ requires at least 52 groups for 5\% and 133 for 2\%. The separate SCORPION gate evaluates eight reserved validation slides using one region-control summary and a training-source prediction threshold, without the two-compartment disagreement test.

\section{Controlled synthetic results}
\label{app:synthetic}

The generator uses 48-dimensional embeddings, a rank-six acquisition process, three tissue compartments, and ordinal labels 0--5. Each scenario has eight seeds and 300 patients with a patient-disjoint 70/30 train/test split and 24 tiles per compartment; the 60- and 170-patient scenarios restrict paired calibration while source scaling, prototype banks, and ordinal-head fitting retain all 210 source-training patients. Let $\delta_m$ be the coordinate-wise median full-dimensional lesion target-minus-source difference for non-source scan $m$. Shift RMSE is $\sqrt{M^{-1}\sum_{m=1}^M\|\delta_m-\widehat\delta_m\|_2^2}$, where $\widehat\delta_m$ is the applied correction. Worst-target QWK is the minimum over non-source scanners, computed over all simulated labels 0--5. Grade flips average disagreement over all unordered scanner pairs and all test slides, including the source scanner. Paired inference uses 10,000 seed-level percentile bootstrap replicates and exact two-sided sign-flip tests, with Holm adjustment over three comparisons. In the acquisition-sensitive regime, RMSE reduction and QWK gain each have adjusted $p=0.03125$. Source QWK is unchanged at 0.9976.

\begin{table}[ht]
\centering\tablefont
\setlength{\tabcolsep}{3pt}
\caption{Complete synthetic regimes. Means include all eight seeds; rejection by the outer screen returns the raw embeddings. Cov.: mean correction coverage. Audit: seeds passing the outer screen.}
\begin{tabular*}{\linewidth}{@{\extracolsep{\fill}}lcrrrrrrr@{}}
\toprule &&\multicolumn{2}{c}{Shift RMSE}&\multicolumn{2}{c}{Worst QWK}&\multicolumn{2}{c}{Grade flip}&\\
\cmidrule(lr){3-4}\cmidrule(lr){5-6}\cmidrule(lr){7-8}
Regime&Audit&Raw&Corrected&Raw&Corrected&Raw&Corrected&Cov.\\\midrule
Acquisition-sensitive&7/8&4.014&1.894&.828&.949&.567&.157&.804\\
Already robust head&7/8&4.031&1.893&.983&.993&.098&.038&.807\\
Unseen scanner&7/8&4.449&2.513&.790&.932&.617&.185&.753\\
Unseen, 1.5$\times$ range&7/8&6.673&5.925&.643&.729&.714&.479&.365\\
Matched tissue response&7/8&3.504&1.630&.865&.961&.555&.150&.813\\
Strong response gap&6/8&5.371&2.985&.739&.908&.624&.275&.686\\
60 calibration patients&0/8&4.014&4.014&.828&.828&.567&.567&.000\\
170 calibration patients&5/8&4.014&2.535&.828&.911&.567&.289&.565\\\bottomrule
\end{tabular*}
\end{table}

With matched tissue response, direct and learned ungated transfer have similar RMSE, 0.964 and 0.962. With a strong response gap, their RMSEs are 3.766 and 1.662. In the acquisition-sensitive regime, learned transfer improves on direct subtraction by 0.794 (95\% CI 0.627--1.002; adjusted $p=0.0234$). This isolates the benefit of learning tissue-dependent transfer in the controlled generator.

\begin{table}[ht]
\centering\tablefont
\caption{Synthetic stress tests in the acquisition-sensitive regime. Material worsening exceeds 2\% of the corresponding clean test regime's median raw non-source shift norm. Accepted worsening is conditional on acceptance and undefined at zero coverage. Means of conditional rates omit seeds with undefined rates.}
\begin{tabular*}{\linewidth}{@{\extracolsep{\fill}}lrrr@{}}
\toprule Stress&Ungated worsening&Coverage&Accepted worsening\\\midrule
One-control conflict&.738&.000&--\\
Concordant control drift&.964&.037&.990\\
Morphology out of support&.003&.000&--\\
Novel acquisition direction&.001&.000&--\\
Insufficient controls&.000&.000&--\\
25\% selector contamination&.001&.415&.000\\
Wrong-slide controls&.697&.804&.692\\
Lesion-response mismatch&.266&.804&.272\\\bottomrule
\end{tabular*}
\end{table}

\end{document}